\documentclass[letterpaper, 10 pt, conference]{ieeeconf}

\IEEEoverridecommandlockouts 

\title{\LARGE \bf
PoI: A Filter to Extract Pixel of Interest from Novel Views for Scene Coordinate Regression
}

\author{Feifei Li, Qi Song, Chi Zhang, Hui Shuai, Rui Huang
}

\usepackage{amsmath,amsfonts}
\usepackage{algorithmic}
\usepackage{algorithm}
\usepackage{array}
\usepackage[caption=false,font=normalsize,labelfont=sf,textfont=sf]{subfig}
\usepackage{textcomp}
\usepackage{stfloats}
\usepackage{url}
\usepackage{verbatim}
\usepackage{graphicx}
\usepackage{cite}
\usepackage{xcolor}

\usepackage{multirow}
\usepackage{arydshln}
\usepackage{makecell}
\usepackage{gensymb}

\begin{document}

\maketitle
\thispagestyle{empty}
\pagestyle{empty}

\begin{abstract}

Neural View Synthesis (NVS) techniques such as NeRF and 3D Gaussian Splatting (3DGS) have enabled photorealistic rendering from novel viewpoints and are increasingly used to augment training data for visual localization. However, these methods fundamentally rely on observed geometry and radiance; they interpolate existing information but cannot hallucinate unseen 3D structures or recover missing content under sparse or extreme viewpoints. As a result, rendered views often exhibit blur, structural distortion, or incomplete geometry. While such imperfections may be tolerated by Camera Pose Regression (CPR) methods, they severely degrade Scene Coordinate Regression (SCR), which requires accurate per-pixel 3D supervision.
To address this limitation, we introduce PoI (Pixel-of-Interest), a framework that enables effective NVS augmentation for SCR-based localization. We first employ 3DGS to render novel views and leverage a single-step diffusion model to refine them, allowing the synthesis of structurally plausible details beyond purely geometry-driven interpolation. However, even diffusion-refined views may contain unreliable pixels. Therefore, we propose a progressive pixel-level filtering strategy based on reprojection error to selectively retain trustworthy synthetic pixels during training while suppressing harmful ones.
Extensive experiments on 7Scenes and Cambridge Landmarks demonstrate that our method consistently improves localization accuracy over strong SCR baselines and achieves state-of-the-art performance with competitive training efficiency. Our results reveal that, for SCR, the benefit of novel view augmentation depends not only on generative realism but also on explicit control of pixel-level reliability.

\end{abstract}

\section{Introduction}

\begin{figure*}[t]
\begin{center}
   \includegraphics[width=0.95\linewidth]{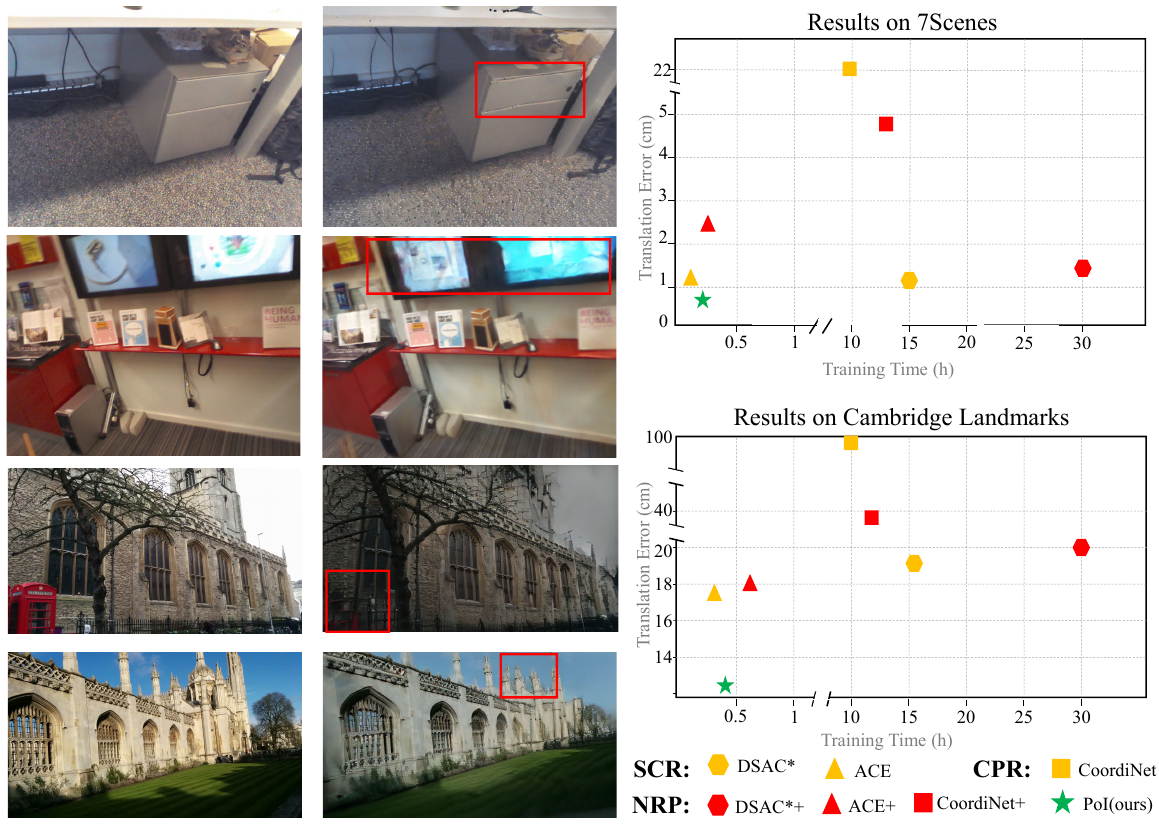}
\end{center}
   \caption{\textbf{Left}: A comparison of query and novel views in the 7Scenes and Cambridge Landmarks datasets highlights notable quality discrepancies. Query frames are typically sharp and structurally consistent, whereas novel frames frequently suffer from blur, missing content, and geometric inconsistencies.
   \textbf{Right}: Translation error versus training time, where `CoodiNet+' denotes using rendered images as query images for CPR method CoodiNet (LENS in this case); `DSAC*+' and `ACE+' denote the method that combines NVS-rendered images and query images as training data for SCR method DSAC* and ACE. `ACE+PoI' denotes our proposed PoI method (ACE-based);
   Analysis reveals that directly adding novel views to the training set will increase training time to some extent, but performance will decrease for the SCR method. On the other hand, our PoI approach can improve the performance with an acceptable time increase.
   }
\label{fig:cover}
\end{figure*}

Visual localization, or camera pose estimation, is a fundamental task in computer vision that involves estimating the 6-degree-of-freedom (6DoF) camera poses within a known scene based on input images. This task plays a crucial role in Simultaneous Localization and Mapping (SLAM)~\cite{izadi2011kinectfusion,mur2015orb,dai2017bundlefusion,tang2018ba-net,lajoie2023swarmslam} and has wide applications in areas such as autonomous driving, robotics, and virtual reality.

Learning-based methods for visual localization can be categorized into two main types: Camera Pose Regression (CPR) methods~\cite{purkait2018synthetic,chen2021direct,ng2021reassessing,taira2018inloc,moreau2022coordinet,moreau2022lens,chen2022dfnet}, which directly regress the camera pose from the input image, and Scene Coordinate Regression (SCR) methods~\cite{brachmann2021visual,brachmann2017dsac,esac,shotton2013scene,valentin2015exploiting,brachmann2023ace}, which first predict dense 3D scene coordinates for image pixels and then estimate the camera pose via 2D–3D correspondences. 
Although SCR methods typically achieve higher localization accuracy due to explicit geometric reasoning, both paradigms require sufficiently dense and well-distributed training data to ensure robust generalization. In practice, collecting large-scale, accurately annotated pose data remains costly and labor-intensive.

To alleviate data scarcity, recent works adopt Neural View Synthesis (NVS) to render synthetic images from novel camera poses, forming so-called Neural Rendering Pose (NRP) estimation methods. For instance, LENS~\cite{moreau2022lens} employs NeRF to render uniformly sampled novel poses and directly treats synthetic images as additional training data. Similarly, DFNet~\cite{chen2022dfnet} utilizes NeRF-W~\cite{nerfw} and incorporates cross-domain feature alignment to mitigate the domain gap between rendered and real images. CaLDiff models camera localization as a generative denoising process over SE(3), iteratively refining noisy pose hypotheses through a learned diffusion network.

However, conventional NVS methods such as NeRF and 3D Gaussian Splatting (3DGS) fundamentally operate through geometric interpolation of observed radiance fields. They cannot hallucinate unseen 3D structures or recover missing content under sparse viewpoints or extreme camera extrapolation. Consequently, rendered views often exhibit blur, incomplete geometry, or structural distortions, as illustrated in the left part of Figure 1. 

While such artifacts may be tolerated in CPR-based NRP methods, their impact differs significantly between CPR and SCR. CPR follows an N-to-1 prediction paradigm, where the pose is regressed from global image features; performance is therefore primarily influenced by overall image realism rather than local pixel fidelity. In contrast, SCR adopts an N-to-N formulation, requiring accurate coordinate prediction for each pixel. Even localized rendering errors can propagate to incorrect 2D–3D correspondences, substantially degrading pose estimation.

As shown in the right part of Figure~\ref{fig:cover}, augmenting SCR training with raw NVS-rendered images not only fails to improve performance but may even degrade accuracy and increase training time. Directly incorporating imperfect synthetic views introduces noisy geometric supervision, which is detrimental to coordinate regression. This explains why existing NRP approaches are largely limited to CPR frameworks, and few studies successfully integrate NVS into SCR-based localization.

To address these limitations, we propose two complementary strategies. First, we incorporate a diffusion-based refinement model to enhance NVS-generated views. Unlike purely geometry-driven rendering, diffusion models possess generative priors that can recover structurally plausible details beyond the observed data distribution, improving the fidelity of synthesized novel views. Nevertheless, even diffusion-refined images may still contain unreliable pixels that violate strict geometric consistency required by SCR.

Second, we introduce a novel Pixel-of-Interest (PoI) module that performs progressive pixel-level filtering during training. Rather than treating rendered images as fully reliable, PoI selectively retains high-confidence pixels based on reprojection consistency while suppressing noisy ones. As training proceeds, unreliable synthetic pixels are gradually excluded, allowing the model to exploit useful novel-view information without compromising geometric supervision.

The main contributions of our work are summarized as follows:
\begin{itemize}
\item We introduce PoI, a pixel-level filtering framework that enables effective integration of NVS into SCR-based localization by removing low-quality rendered pixels.
\item We incorporate diffusion-based refinement into the NVS pipeline, allowing recovery of structurally plausible content beyond geometry-based interpolation.
\item We conduct extensive evaluations on indoor and outdoor datasets, demonstrating that our method achieves state-of-the-art localization accuracy with competitive training efficiency.
\end{itemize}

\section{Related Work}

\textbf{Camera Pose Regression} The CPR methods, i.e., to regress the camera pose from the given image directly, are the most naive ideas in learning-based methods~\cite{kendall2015posenet,brachmann2016uncertainty,brahmbhatt2017mapnet,melekhov2017image,radwan2018vlocnet++,wang2020atloc,hu2020dasgil,arnold2022map,chen2022dfnet,shavit2022camera}. The most straightforward method implicitly uses CNN layers or MLP to represent the image-to-pose correspondence. PoseNet~\cite{kendall2015posenet} first proposes CPR using pre-trained GoogLeNet as the feature extractor. Then, several works focus on improving CPR through additional modules. Geomapnet~\cite{brahmbhatt2017mapnet} estimates the absolute camera poses and the relative poses between adjacent frames. AtLoc~\cite{wang2020atloc} uses a self-attention module to extract salient features from the image. Vlocnet++~\cite{radwan2018vlocnet++} adds a semantic module to solve the dynamic scene and improve the robustness for blockings and blurs. 
Marepo~\cite{chen2024marepo} first regresses the scene-specific geometry from the input images and then estimates the camera pose using a scene-agnostic transformer. 
The CPR method has achieved excellent efficiency and framework simplification, but there is still room for improvement in accuracy. 
\textbf{Scene Coordinate Regression} The SCR-based method~\cite{shotton2013scene,brachmann2017dsac,brachmann2018learning,esac,massiceti2017random,li2018scene,brachmann2021visual,liu2025egfs} aims to estimate the camera pose in two steps: first, estimate the coordinates of the points in the 3D scene corresponding to the 2D pixels, and then estimate the camera pose based on the 2D-3D correspondence relationship.
SCR was initially proposed using the random forest for RGB-D images~\cite{shotton2013scene}. Recently, estimating scene coordinates through RGB input has been widely studied. ForestNet\cite{massiceti2017random} compares the benefits of Random Forest (RF) and Neural Networks in evaluating the scene coordinate and camera poses. ForestNet also proposes a novel method to initiate the neural network from an RF. DSAC~\cite{brachmann2017dsac}, DSAC++~\cite{brachmann2018learning} devise a differentiable RANSAC, and thus the SCR method can be trained in an end-to-end manner. ESAC~\cite{esac} uses a mixture of expert models (i.e., a gating network) to decide which domain the query belongs to, and then the complex SCR task can be split into simpler ones. DSAC*~\cite{brachmann2021visual} extends previous works to applications using RGB-D or RGB images, with or without 3D models. 
This means that in the minimal case, only RGB images will be used as the input to DSAC*, just like most CPR methods. 
More information about the 3D structure will be utilized for most SCR methods than for CPR ones. However, approaches like DSAC* can achieve more accurate estimations even if the input is the same as the CPR method.
ACE~\cite{brachmann2023ace} and GLACE~\cite{wang2024glace} abandon the time-consuming end-to-end supervision module and shuffle all pixels of the scene to improve training efficiency from the sampled poses $P_{novel}$ using additional 3D geometry information and achieve comparable accuracy compared to former methods.

\textbf{Neural Render Pose Estimation} 
In many cases, collecting more high-quality and accurately annotated training data to improve localization accuracy is an effective solution.
To reduce the cost and improve efficiency,
Some works also try to use more flexible NVS to render synthetic images instead of collecting extra data~\cite{chen2021direct,ng2021reassessing,purkait2018synthetic,taira2018inloc,moreau2022lens,chen2022dfnet}, where NVS is a set of techniques to render synthetic views from existing views.
INeRF~\cite{yen2021iNeRF} applies an inverted NeRF to optimize the estimated pose through the color residual between rendered and observed images. However, the initially estimated poses are crucial in guaranteeing the convergence of outputs.
LENS~\cite{moreau2022lens} samples the poses uniformly throughout the area and trains a NeRF-W~\cite{nerfw} to render the synthetic images. Then, the rendered images and poses work as additional training data for the pose regression network. The limitation of LENS lies in the costly offline computation for dense samples. DFNet~\cite{chen2022dfnet} uses direct feature matching between observed and synthetic images generated by histogram-assisted NeRF. The feature match approach is proposed to extract observed or generated images' cross-domain information. All these methods combine the NVS module and the CPR module to optimize the performance of the absolute pose estimation of the photos.

However, since N-to-N prediction of SCR requires higher image fidelity than N-to-1 prediction of CPR, there is a lack of methods to effectively utilize the NVS-generated data in SCR-based methods. 
In this paper, we propose to use NVS-generated data in end-to-end SCR approaches such as ACE and GLACE.
First, we design novel pose sampling methods to meet the multiple requirements of different datasets. To address the problem of varying lighting conditions, we use exposure histogram-assisted 3DGS as the baseline to sample new views of multiple exposures for each sampled pose in outdoor datasets.
Second, we propose a pixel filter to remove bad pixels in rendered images and use query images and remaining rendered pixels to improve the estimation.

\begin{figure*}[t]
\begin{center}
   \includegraphics[width=1.\linewidth]{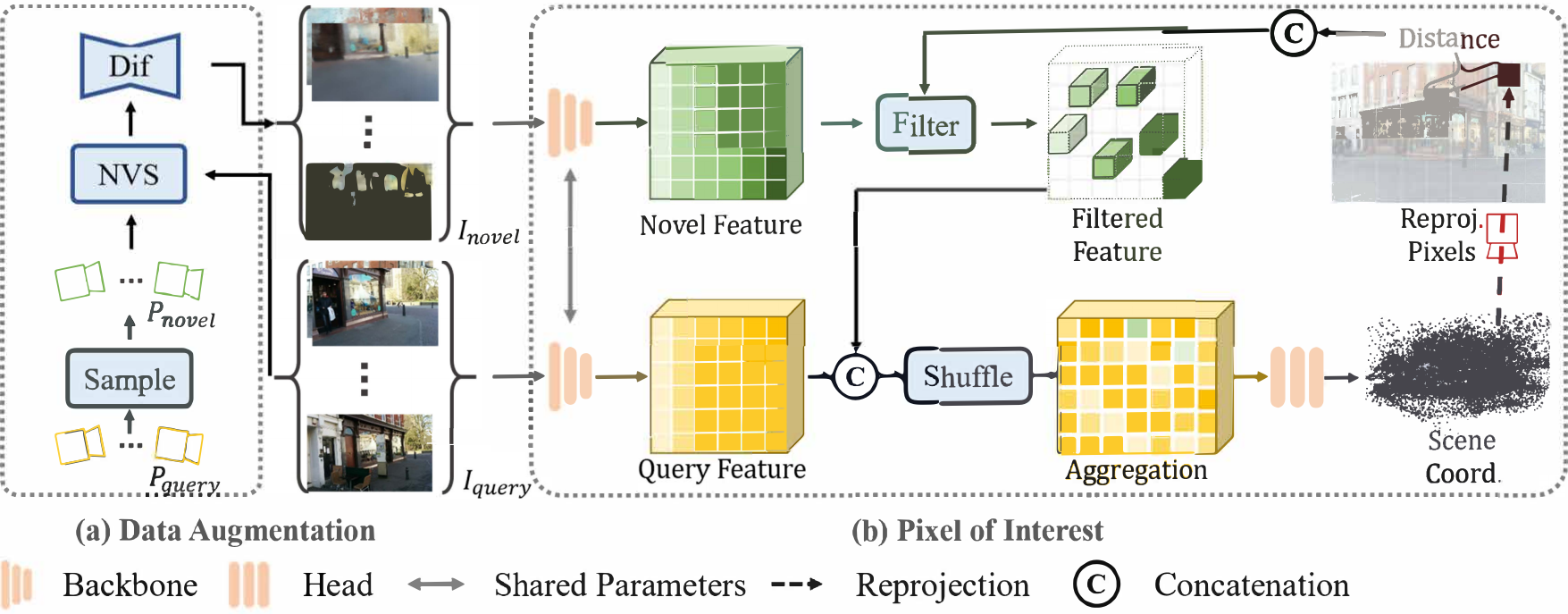}
\end{center}
   \caption{Pipeline of our proposed methods: \textbf{(a) Data Augmentation}: We first sample a group of synthesized camera poses $P_{novel}$ according to the query training pose $P_{query}$ using Fisher Sample. Then, we render the synthesized views $I_{novel}$ from the sampled poses $P_{novel}$ using the novel view synthesis model. 
   \textbf{(b) Architecture of PoI module}: First, a pre-trained scene-irrelevant backbone is applied to extract the features of the input query photos $I_{query}$ and the synthesized novel images $I_{novel}$. Then, the filter is applied to the rendered image features and extracts the features of interest. After that, we combine the query features with the retained features of the novel views and shuffle the pixel-aligned features to get the aggregation. Finally, we estimate the scene coordinates of the pixels using a scene-specific Head. The filtering algorithm is designed based on the reprojection error and the gradient of the estimated scene coordinates.}
\label{fig:pipeline}
\end{figure*}

\section{Method}

In general, the pipeline of our proposed method is shown in Figure~\ref{fig:pipeline}. For the input query images $I_{query}$ and corresponding camera poses $P_{query}$, we first sample the novel camera poses $P_{novel}$. Then we render novel views $I_{novel}$ through 3DGS and diffusion model. Finally, we use the PoI module to estimate the scene coordinates according to the input $I_{query}, I_{novel}$. During the evaluation stage, we use PNP-based RANSAC to infer the camera poses from the scene coordinates.

The following part of this section is arranged as follows:
We first introduce the building blocks of this paper, including NVS models, the diffusion model, the synthetic pose sampling approach, and the baseline of the SCR regressor in Section~\ref{method:blocks};
We introduce the data augment details in Section~\ref{method:data_augment}.
We elaborate on the architecture of the PoI method and introduce the details of the filtering strategy used in the PoI method in Section~\ref{method:poi} and Section~\ref{method:filter};

\subsection{Building Blocks}

\label{method:blocks}
\textbf{Fisher Sample Method} Before PoI training, we would sample novel camera poses $P_{novel}$ and render the corresponding images $I_{novel}$ based on the query images $I_{query}$ and the corresponding camera poses $P_{query}$. 
Since rendered novel views are expected to contain more information about the scenes beyond the training data, we choose to use the Fisher information selection method from FisherRF~\cite{jiang2024fisherrf}.

\textbf{NVS:} This paper uses 3DGS~\cite{kerbl20233dgs} as the baseline for novel view synthesis. A major challenge for 3DGS in outdoor scenarios is the failure caused by changes in lighting conditions, which also exists in the localization problem. To tackle this problem, we apply the luminance histogram method from DFNet~\cite{chen2022dfnet} to adjust the appearance of the rendering images in 3DGS. We generate the exposure embedding from the luminance histogram and output the affine transformation from an MLP used in wild Gaussians~\cite{kulhanek2024wildgaussians}.

\textbf{Diffusion Model: } We use DIFIX3D+~\cite{wu2025difix3d+} in this paper for refinement. DIFIX3D+ is a refinement framework that enhances coarse 3D reconstructions using a `single-step diffusion model' instead of traditional multi-step denoising. Fast 3D pipelines such as sparse-view NeRF or 3D Gaussian Splatting often produce noisy geometry, blurred textures, and view inconsistencies; while diffusion models can correct these issues, iterative sampling is computationally expensive. DIFIX3D+ distills the diffusion prior into a one-pass denoiser that directly maps rendered images from a coarse 3D model to refined outputs.
At inference, the system renders views from a coarse model, refines them in a single forward pass, and uses the improved images to update the 3D representation. The result is sharper appearance, cleaner geometry, and improved consistency with minimal computational overhead.

\textbf{Backbone:} To expedite convergence, we use the pseudo-depth supervision mechanism from ACE for suboptimal pixels exhibiting elevated reprojection errors. Specifically, when a pixel's reprojection error surpasses a predefined threshold, we initially enforce its convergence toward designated pseudo-depth targets through constrained optimization. After a certain epoch, when the estimated values of these pixels fall within the valid threshold, we use conventional supervision for further optimization.

\subsection{Data Augment}
\label{method:data_augment}
\begin{figure}[t]
\begin{center}
   \includegraphics[width=1.\linewidth]{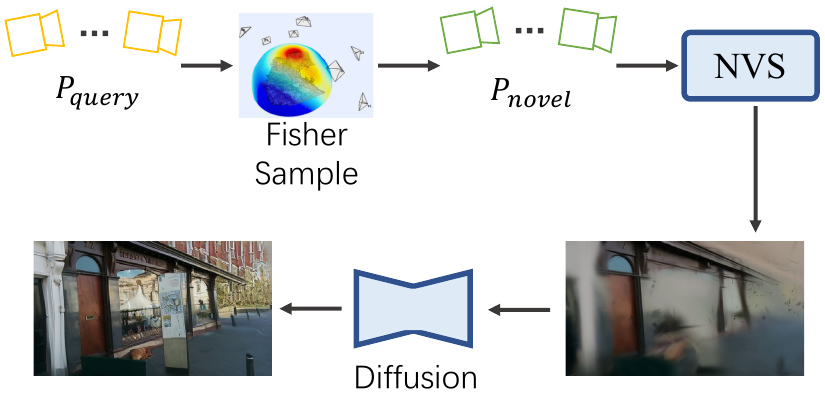}
\end{center}
   \caption{The data augment process under sparse input circumstances.}
\label{fig:augment_sparse}
\end{figure}

We adopt two different data augmentation strategies according to the characteristics of the dataset. The augmentation focuses primarily on how to sample novel camera poses.
When the input data are densely captured—characterized by small inter-frame motion—we generate novel poses using a nearest-neighbor search strategy starting from the evaluation set. Specifically, poses are sampled by identifying spatially adjacent viewpoints in pose space to enrich local coverage.
In contrast, when the input data are relatively sparse, we employ a Fisher sampling strategy to select additional poses, ensuring that the training set reaches a predefined target size. This approach promotes a more uniform and statistically representative distribution of viewpoints across the pose manifold.
As shown in
Figure~\ref{fig:augment_sparse}, the sparse cases take query views along with the camera poses as the input data. Then sample the next best view using the Fisher Sample method to extend the views upto $N$ frames. We use a 3DGS to render coarse novel views and then refine the coarse poses using a diffusion model to remove artifacts and complete the blurred and occluded areas.
 
\subsection{Architecture of PoI}

\label{method:poi}
Current approaches employing novel images as auxiliary inputs for camera pose estimation indiscriminately process entire images, neglecting per-pixel rendering quality variations. This oversight induces dual inefficiencies: (1) Redundant low-fidelity pixel processing will incur significant computational overhead, and (2) Indiscriminate adoption of generated images will reduce the robustness of training.
To address these limitations, we propose a discriminative filtering mechanism that selectively retains high-confidence rendered pixels. Our solution stems from the analysis of the reconstruction properties of 3D Gaussian distribution (3DGS): 3DGS lacks explicit inter-pixel dependency modeling; spatial correlations arise only implicitly through overlapping Gaussian projections. As a result, traditional frame-level filtering can lead to a catastrophic loss of high-fidelity pixels due to the elimination of entire images.
Our methodological innovation establishes a per-pixel filtering paradigm through multi-frame quality consistency analysis. This method effectively decouples the pixel selection decision from the integrity constraints of a single frame.

The architectural framework of PoI is illustrated in Figure~\ref{fig:pipeline}(b). 
The query image $I_{query}$ and rendered novel view $I_{novel}$ are processed through the backbone to yield corresponding query view features $F_{query}$ and novel view features $F_{novel}$. 
Subsequently, the proposed filter is implemented to obtain features of interest (FoI) within $F_{novel}$. These FoI descriptors are then fused with $F_{query}$ through feature concatenation, enabling precise regression of scene coordinates for the corresponding pixels via a coordinate prediction head.

\subsection{Filtering Strategy in PoI}

\label{method:filter}
The filtering pipeline operates through two coordinated mechanisms: \textbf{Subsampling} and \textbf{Reprojection} Co-Filter method.

After data preparation, the amount of rendering data is the same as that of query data. 
During training, we implement the two-step progressive synthetic pixels filtering method to prevent model destabilization from imperfect renderings. 
First, our subsampling protocol extracts features from shuffled synthetic data with Bernoulli probability $p=0.5$.

After that, per-iteration filtration employs a dual-criterion gate function:
\begin{equation}
    \mathcal{G}(x,y) = \mathbb{I}\left[\mathcal{L}_{\text{reproj}} < \tau_r\right]
\end{equation}
where $\tau_r$ is the pre-set threshold of projection distance.

We use reprojection error (the distance between pixel coordinates GT and the re-projected 2D pixel coordinates estimation).

\begin{figure}[t]
\begin{center}
   \includegraphics[width=1.\linewidth]{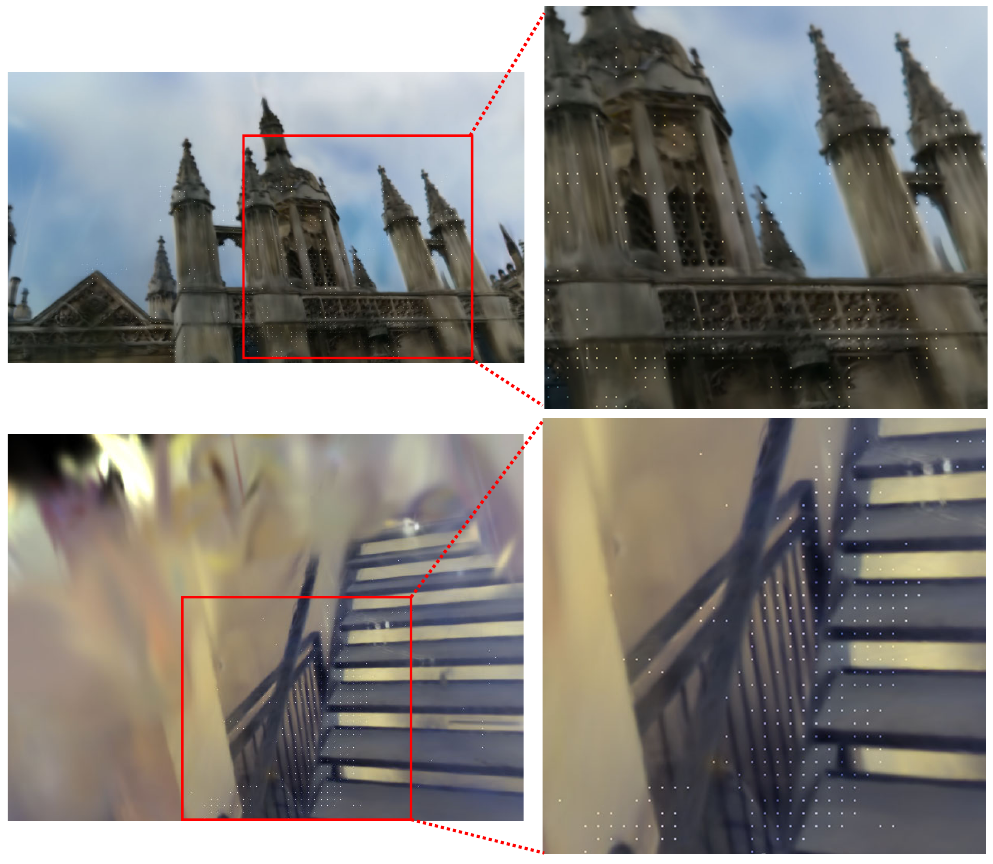}
\end{center}
   \caption{An example of the results of PoI method in dataset 7Scenes and Cambridge Landmarks. To highlight the determined pixels of interest, we scale up the `Value' (V) of the HSV representation of the images.}
\label{fig:poi}
\end{figure}

The filter will remove novel views features of outlier prediction. The remaining features are FoI, and the corresponding pixels of FoI are PoI. Figure~\ref{fig:poi} shows an example of the PoI results in the 7Scenes and Cambridge Landmarks datasets. 
After filtering, we combine and shuffle the features $F$ and FoI and put them into the scene-specific MLP Head to estimate the scene coordinates.
Each pixel is saved with its corresponding frame ID and pixel coordinates to ensure realignment after shuffling.

It is worth mentioning that we have set a dynamic weight for the loss of rendered pixels. Because in the early stage of training, we want the model to converge quickly. After determining the PoI, we gradually reduce the weight of the loss of PoI from 1 to 0.01, while for the pixels from query images, we set the weight to 1 during the entire training process. The loss function is designed as follows:
\begin{equation}
\begin{split}
    \mathcal{L} &= \begin{cases}  
        \mathcal{L}_{rep}^{query}(i), & \text{if } i \in T \\  
         \tilde{\omega} \times \mathcal{L}_{rep}^{poi}(i),  & \text{if } i \in PoI
    \end{cases} \\
    \tilde{\omega} &= \omega_{max} - \frac{I_{iter}}{N_{iter}}(\omega_{max}-\omega_{min})
\end{split}
\end{equation}
where $\mathcal{L}_{rep}$ means the reprojection loss, T denotes query data, $\tilde{\omega}$ denotes the dynamic weight of PoI loss, changing from $\omega_{max}$ (set to 1) to $\omega_{min}$ (set to 0.01). $I_{iter}$ denotes the current iteration number and $N_{iter}$ is the total iterations. All rendered pixels are initially set as PoI. As the training progresses, we rule out outlier prediction points from PoI. At the end of the training, the choice of PoI and the loss weight of PoI are fixed.

\section{Experiment}

\renewcommand{\tabcolsep}{8pt}
\begin{table*}[!t]\footnotesize
	\renewcommand{\arraystretch}{1.2}
	\centering
	\caption{Median errors of camera pose regression methods and scene coordinate regression methods on the 7Scenes dataset~\cite{shotton2013scene}. We \textbf{bold} the best results.}
    
	\label{tab:7scenes}
        \begin{tabular}{c|l|ccccccc|c}
		\cline{1-10}
            &\multirow{2}{*}{Method} & \multicolumn{7}{c|}{Scenes} & \multirow{2}{*}{\makecell{Avg.\\(cm/\degree)}} \\
            && Chess & Fire & Heads & Office & Pumpkin & Kitchen & Stairs &  \\
            \cline{1-10}
            
            \multirow{2}{*}{\rotatebox{90}{CPR}} 
            &PoseNet15  & 10/4.02 & 27/10.0 & 18/13.0 & 17/5.97 & 19/4.67 & 22/5.91 & 35/10.5 & 21/7.74 \\

            
            
            & Marepo & 1.9/0.83 & 2.3/0.92 &  2,1/1.24 & 2.9/0.93 & 2.5/0.88 & 2.9/0.98 & 5.9/1.48 & 2.9/1.04 \\
            
            \cline{1-10}
            
            \multirow{3}{*}{\rotatebox{90}{SCR}} 
            & DSAC* & 0.5/0.17 & 0.8/0.28 & 0.5/0.34 & 1.2/0.34 & 1.2/0.28 & 0.7/0.21 & 2.7/0.78 & 1.1/0.34 \\
            
            & ACE & 0.5/0.18 & 0.8/0.33 & 0.5/0.33 & 1.0/0.29 & 1.0/0.22 & 0.8/0.2 & 2.9/0.81 & 1.1/0.34 \\

             & GLACE & 0.6/0.18 & 0.9/0.34 & 0.6/0.34 &  1.1/0.29 &  0.9/0.23 & 0.8/0.20 & 3.2/0.93 & 1.2/0.36 \\
            
            \cline{1-10}
            
            \multirow{5}{*}{\rotatebox{90}{\makecell{NRP}}}

            & LENS & 3/1.30 & 10/3.70 &  7/5.80 & 7/1.90 & 8/2.20 & 9/2.20 & 14/3.60 & 8/3.00 \\

            & DFNet & 3/1.12 & 6/2.30 & 4/2.29 & 6/1.54 & 7/1.92 & 7/1.74 & 12/2.63 & 6/1.93 \\

            & GSplatLoc & 0.43/0.16 & 1.03/0.32 & 1.06/0.62 &  1.85/0.4 &  1.8/0.35 & 2.71/0.55 & 8.83/2.34 & 2.53/0.68 \\

            & PoI(ours) & 0.4/0.13 & \textbf{0.5}/0.27 &  0.4/\textbf{0.15} & 0.7/0.2 & 0.8/0.17 & \textbf{0.5/0.2} & 2.2/\textbf{0.55} & 0.8/0.25 \\
            
            & GLPoI(ours) & \textbf{0.3/0.10} & 0.6/\textbf{0.21} &  \textbf{0.3}/0.22 & \textbf{0.4/0.19} & \textbf{0.7/0.13} & \textbf{0.5/0.2} & \textbf{2.1}/0.62 & \textbf{0.7/0.24} \\
            
            \cline{1-10}	

	\end{tabular}
\end{table*}

\subsection{Experiments Setup}

\renewcommand{\tabcolsep}{4pt}
\begin{table*}[!t]\footnotesize
	\renewcommand{\arraystretch}{1.2}
	\centering
	\caption{Results on Cambridge Landmarks, because of the obvious gap between SCR-based methods and CPR-based methods, we only list SCR-based methods. Column `Mapping time' shows the training time of these methods (the GS model can be trained offline, and the training time of the GS model is shorter than PoI method), and column `Mapping size' is the memory consumption for saving the parameters of the network. We \textbf{bold} the best result for group `SCR' and group `NRP' separately.}
	\label{tab:cambri}
	\begin{tabular}{c|l|c|c|c|ccccc|c}
		\cline{1-11}
            & \multirow{2}{*}{Method} & \multirow{2}{*}{\makecell{Mapping with \\ Depth/Mesh}} & \multirow{2}{*}{\makecell{Mapping\\Time}} & \multirow{2}{*}{\makecell{Map\\Size}} & \multicolumn{5}{c|}{Scenes} & \multirow{2}{*}{\makecell{Avg.\\(cm/\degree)}}  \\
            &&&&& King's & Hospital & Shop & Church & Court &   \\
            
            
            

            \cline{1-11}
            
            \multirow{6}{*}{\rotatebox{90}{SCR}} 
            
            & SANet~\cite{yang2019sanet} & Yes & ~1min & ~260M & 32/0.5 & 32/0.5 & 10/0.5 &  16/0.6 & 328/2 & 84/{0.8}\\
            
            & SRC~\cite{dong2022src} & Yes & 2min & 40M & 39/0.7 &  38/0.5 & 19/1 & 31/1.0 & 81/0.5 & 42/0.7\\
            
            & DSAC*~\cite{brachmann2021visual} & No & 15h & 28M & 18/\textbf{0.3} & 21/0.4 & 5/0.3 &  15/0.6 & 34/0.2 & 19/0.4\\
            
            & Poker~\cite{brachmann2023ace} & No & 20min & 16M & 18/0.3 & 25/0.5 &  5/0.3 & 9/0.3 & 28/0.1 & 17/\textbf{0.3} \\

            & EGFS~\cite{liu2025egfs} & No &21min & 9M & \textbf{14/0.3} & 28/\textbf{0.1} &  19/0.4 & \textbf{5/0.2} & \textbf{10}/0.3 & 15/\textbf{0.3} \\

            & GLACE~\cite{wang2024glace} & No & 20min & 13M & 19/\textbf{0.3} &  17/0.4 & \textbf{4/0.2} & 9/0.3 & 19/0.1 & 14/\textbf{0.3} \\            
            
            \cline{1-11}
            
            \multirow{4}{*}{\rotatebox{90}{\makecell{NRP}}} 

            

            & LENS~\cite{moreau2022lens} & No & - & - & 33/0.5 & 44/0.9 &  27/1.6 & 53/1.6 & - & - \\

            & GSplatLoc~\cite{sidorov2024gsplatloc} & No & - & - & 27/0.46 & 20/0.71 &  5/0.36 & 16/0.61 & - & - \\


            & PoI (ours) & No & 25min & 16M & 16/\textbf{0.3} & 17/0.4 &  \textbf{4/0.2} & 8/0.3 & 16/\textbf{0.1} & 12.2/\textbf{0.3} \\
            
            & GLPoI (ours) & No & 25min & 13M & 16/\textbf{0.3} & \textbf{14}/0.4 & \textbf{4/0.2} & 7/0.3 & 16/\textbf{0.1} & \textbf{11.4/0.3} \\
            
            \cline{1-11}	

	\end{tabular}
\end{table*}

The proposed framework is evaluated on two publicly available benchmarks: the Microsoft 7Scenes dataset~\cite{shotton2013scene} (\textbf{with SFM-derived pseudo ground truth}) and the Cambridge Landmarks dataset~\cite{kendall2015posenet}. The methodology operates exclusively on RGB inputs paired with camera extrinsic parameters, excluding depth information from 7Scenes and structural reconstruction priors from Cambridge Landmarks. Full-resolution RGB images are preserved to ensure precise pose estimation capabilities.

To optimize computational efficiency, pose sampling and novel view synthesis are preprocessed offline, with generated camera poses and rendered images stored on disk. These precomputed rendered images are loaded with the original training data during the training phase. A specialized strategy is implemented for the `Kitchen' scene, where training samples are partitioned into two clusters based on camera pose distributions. Following the ACE-inspired protocol, two distinct models are trained on these clusters. At inference time, the final pose estimation is selected from the model demonstrating superior RANSAC consensus through a higher proportion of inlier correspondences.
All experiments utilized NVIDIA V100 GPUs. PoI training employed a single GPU, whereas GLPoI training used four GPUs in a distributed data-parallel setup to expedite convergence.

\subsection{Quantitative Results}

The comparison of median translation and rotation errors between our proposed methods with different camera pose regression methods (top), scene coordinate regression methods (middle), and neural render pose estimation methods (bottom) in dataset 7Scenes is shown in Table~\ref{tab:7scenes}. 
Generally speaking, scene coordinate regression methods outperform absolute camera pose regression methods in both translations and orientations. 
Unlike~\cite{sidorov2024gsplatloc}, we include DFNet and LENS in NRP, although they use offline NVS.
Our proposed method outperforms DSAC* and ACE by exploiting the additional information from the rendered novel views. Our `GLPoI' beats `GLACE' and achieves the state of the art.

The results of the Cambridge Landmarks datasets are shown in Table~\ref{tab:cambri}. Since there are apparent gaps between scene coordinate regression methods and absolute camera pose regression methods, we present only the results for SCR and NRP methods. 
We come to a similar conclusion as that of 7Scenes. 
Although we use additional rendered data, it can achieve training efficiency comparable to ACE. 

\subsection{Ablation of PoI}

To evaluate the effectiveness of PoI, we conduct ablation experiments under multiple configurations, as summarized in Table~\ref{tab:abla_poi}. We first define \textbf{case `base'} as training with only query images from the original training set, following a setting comparable to ACE. \textbf{Case `dif+poa'} denotes training with both query data and all pixels from diffusion-refined novel views, without any filtering.
\textbf{Case `dif+por'} represents a pixel-of-random strategy, where randomly sampled pixels from diffusion-refined NVS are used, while keeping all other settings identical to \textbf{case `dif+poi'}.
\textbf{Case `dif+poi'} corresponds to our full method with reprojection-based pixel filtering.
\textbf{Case `3dgs+poi'} employs 3DGS alone for novel view generation without diffusion refinement. 

The results indicate that directly incorporating rendered images without filtering (\textbf{`dif+poa'}) leads to worse performance than the baseline on 7Scenes. This suggests that low-quality synthetic pixels introduce noisy geometric supervision, which misguides coordinate regression and degrades localization accuracy.

A potential concern is that improving the quality of novel view synthesis might reduce the necessity of PoI, as better renderings could theoretically diminish the impact of unreliable pixels. To investigate this, we further evaluate different NVS backbones.
Experimental results show that combining 3DGS with diffusion produces consistently better performance than 3DGS alone, indicating that higher-quality synthesis enhances the effectiveness of PoI rather than diminishing it.

Nevertheless, even diffusion-refined renderings remain insufficient for direct use in SCR, as pixel-level geometric inaccuracies still exist. This confirms that generative enhancement alone cannot fully satisfy the strict supervision requirements of scene coordinate regression, and explicit pixel-level filtering remains essential.

\renewcommand{\tabcolsep}{2pt}
\begin{table}[!t]\footnotesize
    \renewcommand{\arraystretch}{1.2}
    \centering
    \caption{Median errors of different implementations of PoI method on 7Scenes and Cambridge dataset.}
	\label{tab:abla_poi}
	\begin{tabular}{lccc|ccc}
		\cline{1-7}
            \multirow{2}{*}{Method} & \multicolumn{3}{c|}{7Scenes} & \multicolumn{3}{c}{Cambridge Landmarks}\\
            & trans$\downarrow$ & rot$\downarrow$ & $U_{5\text{cm},5\degree}\uparrow$ & trans$\downarrow$ & rot$\downarrow$ & $U_{10cm,5\degree}\uparrow$ \\
            \cline{1-7}
            base & 1.1cm & 0.3\degree & 97.1\% & 17.7cm & \textbf{0.3}\degree & 32.4\% \\
            dif+poa & 2.3cm & 1.58\degree & 89.6\% & 17.6cm & \textbf{0.3}\degree & 32.2\%\\
            dif+por & 1.4cm & 0.46\degree & 95.9\% & 18.0cm & \textbf{0.3}\degree & 30.4\%\\
            3dgs+poi & 1.0cm & 0.3\degree & 99.1\% & 15.4cm & \textbf{0.3}\degree & 34.7\% \\
            dif+poi & \textbf{0.8cm} & 0.25\degree & \textbf{98.7\%} & \textbf{12.2cm} & \textbf{0.3}\degree &  \textbf{38.7\%} \\

            \cline{1-7}	
	\end{tabular}
\end{table}

\subsection{Sparse-Input Cases}

\renewcommand{\tabcolsep}{3pt}
\begin{table}[!t]\footnotesize
	\renewcommand{\arraystretch}{1.2}
	\centering
	\caption{Median errors of our proposed method with sparse input on 7Scenes and Cambridge dataset.}
	\label{tab:sic}
	\begin{tabular}{lccc|ccc}
		\cline{1-7}
            \multirow{2}{*}	{Method} & \multicolumn{3}{c|}{7Scenes} & \multicolumn{3}{c}{Cambridge Landmarks}\\
            & trans$\downarrow$ & rot$\downarrow$ & $U_{5cm,5\degree}\uparrow$ & trans$\downarrow$ & rot$\downarrow$ & $U_{10cm,5\degree}\uparrow$ \\
            \cline{1-7}
            base &  2.6cm & 0.7\degree & 68.9\% & 435cm & 2.2\degree & 15.7\%\\
            3dgs-poi & 1.6cm & 0.4\degree & 93.0\% & 22.7cm & \textbf{0.3}\degree & 21.9\%\\
            dif-poi & \textbf{1.3cm} & \textbf{0.3}\degree & \textbf{94.7\%} & \textbf{18.3cm} & \textbf{0.3}\degree & \textbf{24.6\%}\\
            
            \cline{1-7}	
            	\end{tabular}
\end{table}

To validate the effectiveness of diffusion-based novel view synthesis (diffusion-NVS) methods under sparse-view settings, we design a controlled experimental protocol on the 7-Scenes dataset. Specifically, we reduce the input training set to only 10 images per scene and train a 3D Gaussian Splatting (3DGS) model under this extremely sparse configuration.

During training, we adopt the Fisher Information-based active view selection strategy proposed in FisherRF to sample additional informative camera poses. Concretely, candidate poses are evaluated using the Expected Information Gain (EIG) criterion derived from the Fisher Information matrix, and the most informative poses are selected and progressively incorporated into the training set. These newly sampled views are then rendered via diffusion-based novel view synthesis and used as pseudo-observations to further optimize the 3DGS model.

Through this iterative expansion process, the training set is gradually enlarged from 10 views to 25 views. The expanded dataset is subsequently used to refine the Point-of-Interest (PoI) reconstruction model, enabling improved geometric consistency and rendering quality under sparse supervision.

The quantitative results are summarized in Table~\ref{tab:sic}, demonstrating that the proposed diffusion-NVS augmented pipeline (case `dif-poi') significantly improves the localization performance of PoI compared to training with the original sparse inputs alone.

\begin{figure}[t]
\begin{center}
   \includegraphics[width=1.\linewidth]{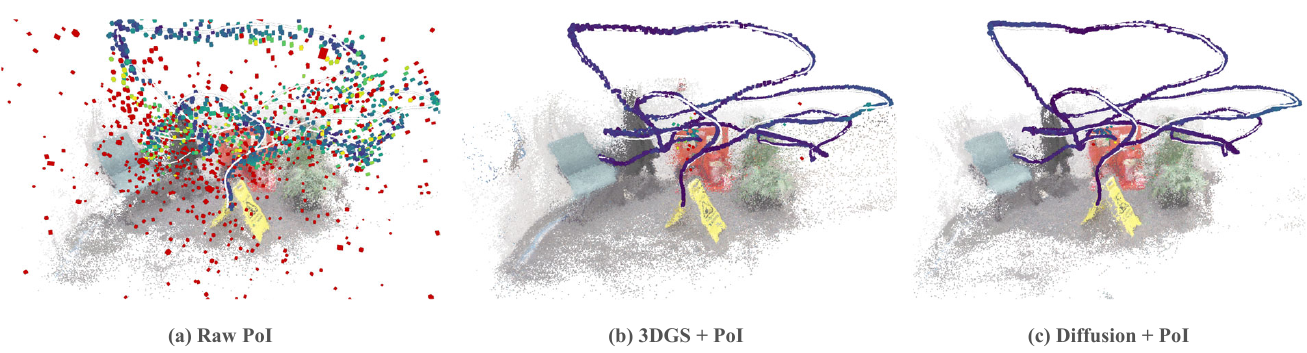}
\end{center}
   \caption{Visualized results of scene coordinates and localization trajectories.}
\label{fig:visualized_sparse}
\end{figure}

We construct the mesh based on the estimated scene coordinates and visualize the camera pose estimation results in Figure~\ref{fig:visualized_sparse}. We may find that for the raw PoI, the pose estimation error is relatively large, and the quality of the reconstructed details is low. For the `NVS + diffusion' case, the performance is much better.

\section{Conclusion}

We proposed PoI, a pixel-level filtering framework that enables effective integration of Neural View Synthesis into Scene Coordinate Regression (SCR)-based visual localization. Unlike CPR methods, SCR requires strict per-pixel geometric consistency and is highly sensitive to rendering artifacts. Geometry-driven NVS methods such as NeRF and 3DGS can only interpolate observed radiance and cannot recover unseen 3D structures, which limits their direct applicability to SCR. 

By incorporating diffusion-based refinement, we enhance the structural plausibility of synthesized novel views beyond purely geometry-based interpolation. To further ensure geometric reliability, PoI selectively retains high-confidence synthetic pixels through a reprojection-based filtering strategy during training. 

Extensive experiments demonstrate that our approach consistently improves localization accuracy and achieves state-of-the-art performance with competitive training efficiency. Our results highlight that successful NVS augmentation for SCR requires both generative enhancement and explicit pixel-level reliability control.

{\small
\bibliographystyle{IEEEtran}
\bibliography{ref}
}

\end{document}